\documentclass{article} 
\usepackage{nips11submit_e,times}
\usepackage{amsmath,amsfonts,amssymb,bbm}
\usepackage{times}
\usepackage{graphicx}
\usepackage{color}
\usepackage{multirow}
 \usepackage{booktabs}
 \usepackage{psfrag}
\usepackage{rotating}
\usepackage{bbm}
\usepackage{latexsym}
\usepackage{url}
\usepackage{caption}
\usepackage{subcaption}
\usepackage{tikz}

\nipsfinaltrue
\title{The Infinite Degree Corrected Stochastic Block Model}

\author{Tue Herlau, Mikkel N. Schmidt, Morten M{\o}rup\\
\{tuhe,mnsc,mmor\}@dtu.dk\\
Section for Cognitive Systems, DTU Compute\\
Technical University of Denmark}

\newcommand{\CRP}{\mathrm{CRP}}

\newcommand{\Dir}{\mathrm{Dirichlet}}

\newcommand{\Poisson}{\mathrm{Poisson}}
\newcommand{\Gam}{\mathrm{Gamma}}
\newcommand{\m}[1]{\boldsymbol{#1}}
\newcommand{\Lfrac}{L_{\text{frac} } }

\begin{document}

\maketitle

\begin{abstract}
In Stochastic blockmodels, which are among the most prominent statistical models for cluster analysis of complex networks, clusters are defined as groups of nodes with statistically similar link probabilities within and between groups.
A recent extension by Karrer and Newman incorporates a node degree correction to model degree heterogeneity within each group. Although this demonstrably leads to better performance on several networks it is not obvious whether
modelling node degree is always appropriate or necessary.
We formulate the degree corrected stochastic blockmodel as a non-parametric Bayesian model, incorporating a parameter to control the amount of degree correction which can then be inferred from data. Additionally, our formulation yields principled ways of inferring the number of groups as well as predicting missing links in the network which can be used to quantify the model's predictive performance.
On synthetic data we demonstrate that including the degree correction yields better performance both on recovering the true group structure and predicting missing links when degree heterogeneity is present, whereas performance is on par for data with no degree heterogeneity within clusters. On seven real networks (with no ground truth group structure available) we show that predictive performance is about equal whether or not degree correction is included; however, for some networks significantly fewer clusters are discovered when correcting for degree indicating that the data can be more compactly explained by clusters of heterogenous degree nodes.


\end{abstract}


\section{Introduction}
The stochastic blockmodel (SBM) \cite{white1976social,holland1983,nowicki2001estimation} has become a prominent tool for modeling group structure in complex networks~\cite{Guimera14122009}. However, as pointed out by Karrer and Newman \cite{karrer2011stochastic}, the stochastic blockmodel has a tendency to group nodes according to their degree such that high degree nodes group together even though their patterns of interactions with the remaining network may differ. This grouping thus reflects aspects of node degree rather than overall statistical patterns in the network. To alleviate this issue, Karrer and Newman introduced the degree corrected stochastic blockmodel (DCSBM) \cite{karrer2011stochastic}. In their model, additional parameters modeling node degree heterogeneity are introduced allowing nodes of varying degree to be clustered together, and they demonstrate that including this degree correction reduces the tendency to group nodes according to their degree distribution~\cite{karrer2011stochastic}. The parameters in the DCSBM model are inferred using maximum likelihood (ML) estimation and since closed form expressions for the ML estimates of the additional degree correction parameters are available, the computational complexity of the inference procedure is similar to inference in the SBM.

Although Karrer and Newman demonstrate on several network datasets that degree correction leads to better performance~\cite{karrer2011stochastic}, it is not obvious whether including a degree correction is always appropriate on real network data. Furthermore, the number of groups used in the analysis is likely to influence the results since groups of heterogenous node degree can be reasonably modelled by a number of homogenous subgroups. Not handling this issue in a principled manner could potentially confound the results. Finally, an important subject of network modelling is validation. Although many real networks are hypothesized to possess group structure, no ground truth clustering is available which makes it difficult to assess the goodness of the obtained clustering. A popular alternative is to measure the predictive performance on held out links in the network. In order to do this in a principled manner the methods must be able to handle missing entries in the network data as well as define a predictive distribution over the missing entries.


In this paper we address these three important challenges when modeling network data by the DCSBM:
\begin{itemize}
\item Can we infer the extent in which degree correction is necessary?
\item How can we determine the number of components?
\item How can we predict links in the DCSBM?
\end{itemize}
In particular, we formulate a non-parametric Bayesian generative model for the DCSBM. The number of components are inferred using the Chinese Restaurant Process which has previously been used to determine the number of components in stochastic blockmodels~\cite{kemp2006learning,xu2006learning}. Our generative model is characterized by admitting a simple inference procedure in which both the degree parameter and group interactions can be analytically marginalized out such that inference reduces to estimating the assignments of nodes to clusters as for the DCSBM. We address the link-prediction problem using Markov chain Monte Carlo (MCMC) imputation. By infering the hyper-parameter in the prior distribution of the parameters that account for heterogenous node degree our model is able to learn the extent to which a degree correction is necessary, possibly reducing to an uncorrected stochastic blockmodel. On synthetic as well as seven real networks we demonstrate the utility of our proposed model for determining the number of components, link-prediction, and inferring the magnitude of the parameter controlling degree correction.

Past work on the SBM and DCSBM has not treated the problem of inferring components, presence of degree heterogeneity and link prediction under one unified framework. Although Bayesian approaches to inferring components and link prediction has a long history for the SBM~\cite{Guimera14122009,kemp2006learning,xu2006learning}, most work on the DCSBM has been focused on other inference methods. As noted, Karrer and Newman~\cite{karrer2011stochastic} treated the problem of inference in the DCSBM from a ML perspective. A related approach was taken by Peixoto~\cite{peixoto2012entropy} who considered degree-correction as constraints on a blockmodel ensemble and derived an entropy-based cost function. 
 For the SBM, model relying on a minimum description length based approach to learning has been proposed for inference giving rise to an efficient maximization procedure~\cite{rosvall2007information}. The MDL approach by Rosvald et al.~\cite{rosvall2008maps} allows degree correction but is otherwise analytically different from the DCSBM. For the DCSBM minimum-description length based procedures was considered by Peixoto~\cite{peixoto2013parsimonious} to give an efficient MCMC-based inference procedure, see also~\cite{peixoto2014efficient} for additional discussion of this approach and an application to the problem of estimating the number of components. The belief propagation method of Decelle et al.~\cite{decelle2011inference,decelle2011asymptotic} may also be applied to the DCSBM. More related to our approach is that of Yan et al.~\cite{yan2012model} who consider the problem of inferring the number of groups in the DCSBM from a model-selection perspective.

While these approaches represent important contributions to the problem of jointly modelling degree heterogeneity and block structure, none of the current proposals are based on a Baysian generative model and allow joint inference of degree-correction, number of components and missing links using a MCMC-based approach.


\section{Methods}
Let $\m A$ be the adjacency matrix of an undirected observed network of $n$ nodes such that $A_{ij}$ is the number of links between node $i$ and $j$. We allow a positive number of self-links $A_{ii}$ in our model definition (note that in the original formulation of DCSBM \cite{karrer2011stochastic} $A_{ii}$ is defined as twice the number of self-links).
The DSCBM model~\cite{karrer2011stochastic} for an undirected graph assumes that the links between nodes $i$ and $j$ follow a Poisson distribution
\begin{equation}
\mbox{for $i\neq j$: }  A_{ij} \sim \Poisson\big(\theta_i\eta_{z_iz_j}\theta_j)\big). \label{eq:1}
\end{equation} 
The parameter $\eta_{\ell m}$ controls the probability of links between nodes in group $\ell$ and $m$, $z_i = \ell$ indicate node $i$ is assigned to group $\ell$ and $\theta_i$ is a node specific parameter that regulates this link probability and thus accounts for heterogenous node degrees. The model is subject to the constraint that $\sum_{i} \delta_{z_i \ell} \theta_i=1$ for all groups $\ell$, i.e. the sum of the $\theta_i$ within each group is one.

We presently propose a non-parametric Bayesian generative model that extends the DCSBM dubbed the Infinite Degree Corrected Stochastic Blockmodel (IDCSBM). Like the DCSBM we also maintain node weights $\theta_i$ to control the degree, however, to arrive at a Bayesian formulation we assume the weights within each group is drawn from a Dirichlet distribution. More precisely, for each group $\ell$ containing $n_\ell$ nodes, we introduce a $n_\ell$-dimensional vector of weights $(\phi_i)_{z_i = \ell}$ drawn from a Dirichlet distribution and define $\theta_i = n_\ell \phi_i$ in eq.~\eqref{eq:1}.

The scaling by $n_\ell$ makes the average degree of any given node independent on the size of the group the node belongs to. The full model now consists of (i) generating a random partition, (ii) generating the interaction between each group of the partition $\eta_{\ell m}$ from a gamma distribution, (iii) for each group, generate $(\phi_i)_{z_i = \ell}$ from a Dirichlet distribution and rescale with $n_\ell$, and finally (iiii) use eq.~\eqref{eq:1} to generate the number of links $A_{ij}$ between node $i \neq j$.

The full model is given generatively below. For analytical convenience the model assumes a particular parametrization of the self-links $A_{ii}$, a point we will return to later. 
\begin{align}
& &   \m z          & \sim \CRP(\alpha),              &  & \textit{cluster assignment,}\\
\mbox{for each $\ell$ } & &  (\phi_i)_{z_i = \ell} & \sim  \Dir(\gamma \boldsymbol{1}_{(n_\ell)})& & \nonumber \\
 & & \theta_i & = n_{z_i} \phi_i, & & \textit{relative node degree,} \label{eq:3b} \\
 \mbox{for $\ell \leq m$ } & &  \eta_{\ell m} & \sim \Gam(\kappa,\lambda),        &  & \textit{link rate,}  \\
 \mbox{for $i < j$ } & &  A_{ij}        & \sim \Poisson(\theta_i\eta_{z_iz_j}\theta_j), &  & \textit{link weight,} \label{eq:5}\\
 \mbox{for $i = j$}  & & A_{ii} & \sim \Poisson\left(\frac{1}{2}\theta_i^2\eta_{z_iz_i}\right). &  & \nonumber
\end{align}
In the above $\boldsymbol{1}_{(n_\ell)}$ is a vector of ones with length $n_\ell$, $N=\sum_{\ell=1}^L n_\ell$ is the total number of nodes and $L$ is the number of groups. As a prior over the node partitions $\m z$ we use the Chinese Restaurant Process (CRP) parameterized by a single parameter $\alpha$ controlling the distribution of group size~\cite{aldous1985exchangeability}. A potential advantage of the CRP over for instance a uniform prior over partitions is the CRP is consistent under projections whereas the uniform prior is not. The simplest example is the case where $\m z$ is a partition of two nodes assigned to the same group (i.e. $z_1 = z_2 = 1$) and we consider a partition obtained by including a third node. In this case for the CRP it holds: $p(z_1 = z_2=1 |\alpha) = p(z_1=z_2=1,z_3=1|\alpha)+p(z_1=z_2=1,z_3=2|\alpha)$, however for the uniform prior the left-hand side is $\frac{1}{2}$ and the right-hand side $\frac{2}{5}$. 

Notice the role played by $\gamma$ in the Dirichlet distribution in eq.~\eqref{eq:3b}. If $\gamma \rightarrow \infty$, we will have $\phi_i \rightarrow \frac{1}{n_\ell}$ for $z_i = \ell$ or simply $\theta_i \rightarrow 1$ for all $i$ (the limits are understood in distribution) and the model is thus independent of degree in eq.~\eqref{eq:1}. On the other hand, for $\gamma \rightarrow 0$, within each group $\ell$ a single node, $i^*$, will have mass $\theta_{i^*} = n_{\ell}$ and the network become very nearly entirely dominated by a few greedy nodes. We return to the properties of the model in section \ref{sec:properties}. The advantage of a Bayesian formulation is that we can not only infer $\theta_i$, but also a distribution of the degree-correction variable $\gamma$ representing the appropriateness of modelling degree heterogeneity for the network.

Collecting variables of the same type the joint density factorizes as:
\begin{align}
p(\boldsymbol{A},\boldsymbol{\phi},\boldsymbol{\eta},\boldsymbol{z}|\alpha,\gamma,\kappa,\lambda) & =
p(\boldsymbol{A}|\boldsymbol{\theta},\boldsymbol{\eta},\boldsymbol{z})
p(\boldsymbol{\eta}|\kappa,\lambda)
p(\boldsymbol{\phi}|\boldsymbol{z},\gamma)p(\boldsymbol{z}|\alpha). \label{eq:9}
\end{align}
The model thus depend on parameters $(\alpha,\gamma,\kappa,\lambda)$. While one could fix these at a particular value, a more principal approach we have taken is to introduce vague non-informative priors and sample these as well~\cite{Jaynes2003}. Either choice has no effect on the following derivation below. In our notation the relevant densities are
\begin{align}
p(\m z|\alpha) & = \frac{\alpha^L\Gamma(\alpha)}{\Gamma(N+\alpha)}\prod_{\ell=1}^L\Gamma(n_\ell) & & \textit{(Chinese retaurant process)}, \label{eq:7}\\
\Dir(\boldsymbol x | \boldsymbol \gamma) & = \frac{1}{B(\boldsymbol \gamma)} \prod_i x_i^{\gamma_i-1},  &  B(\boldsymbol \gamma) & = \frac{\prod_i \Gamma(\gamma_i)}{\Gamma(\sum_i \gamma_i)}, \\
\Gam(x | \kappa, \lambda) & = \frac{1}{G(\kappa,\lambda)} x^{\kappa-1}e^{-\lambda x},
& G(\kappa, \lambda) & = \lambda^{-\kappa}\Gamma(\kappa).
\end{align}
The advantage of the proceeding formulation is the use of the Dirichlet distribution within each group, and the particular parametrization of $A_{ii}$, that allow the node weights as well as group interactions to be integrated out analytically. To see this we introduce the short-hand notation for between and within-group link counts
\begin{align}
N^+_{\ell m} & = \left\{\begin{array}{ll} \sum_{i : z_i = \ell, j : z_j = m} A_{ij} & \ell \neq m \\
\sum_{i \leq j : z_i = z_j = \ell} A_{ij} & \ell = m \end{array},\quad
\right. N_{\ell m}=\left\{\begin{array}{ll}n_{\ell}n_{\ell}/2&\text{if } \ell=m\\
n_{\ell}n_{m}&\text{otherwise}\end{array}.\right.
\end{align}
as well as node degrees $k_i=\sum_{j} A_{ij}$ and $\hat{k}_i = k_i + A_{ii}$. It now follows by some algebra
\begin{align}
p(\boldsymbol{A}|\boldsymbol{\theta},\boldsymbol{\eta},\boldsymbol{z})
& = \prod_{i<j} \frac{(\theta_i\eta_{z_iz_j}\theta_j)^{A_{ij}}}{A_{ij}!}\exp{(-\theta_i\eta_{z_iz_j}\theta_j)}
\prod_{i} \frac{\left(\frac{\theta_i^2\eta_{z_iz_i}}{2}\right)^{A_{ii}}}{A_{ii}!}\exp{(-\frac{1}{2}\theta_i^2\eta_{z_iz_i})}  \nonumber \\
& = \frac{1}{\prod_{i\leq j} A_{ij}!\prod_i 2^{A_{ii}}}\prod_{\ell \leq m} \eta_{\ell m}^{N^+_{\ell m}}
\exp{(-\eta_{\ell m}N_{\ell m})}\prod_i\theta_i^{k_i+A_{ii}}  \nonumber \\
& = \frac{1}{\prod_{i\leq j} A_{ij}!\prod_i 2^{A_{ii}}} \left[\prod_{\ell \leq m} \eta_{\ell m}^{N^+_{\ell m}}
\exp{(-\eta_{\ell m}N_{\ell m})}\right]\left[\prod_\ell n_{\ell}^{\hat{k}_\ell}\prod_{i : z_i=\ell}\phi_i^{\hat{k}_i} \right]  \label{eq:11} \\
p(\boldsymbol{\eta}|\kappa,\lambda)
& = \prod_{\ell\leq m} \frac{1}{G(\kappa,\lambda)}\eta_{\ell m}^{\kappa-1}\exp{(-\eta_{\ell m}\lambda)}   \label{eq:12}\\
p(\boldsymbol{\phi}|\boldsymbol{z},\gamma)
& = \prod_\ell \frac{1}{B(\gamma \boldsymbol{1}_{(n_\ell)})} \prod_{i:z_i=\ell} \phi_i^{\gamma-1} \label{eq:13} 
\end{align}
Inserting 
 into eq.~\eqref{eq:9}, collecting terms and exploiting the conjugacy of the Dirichlet and Gamma distributions to the Poisson distribution we can analytically marginalize (i.e., collapse) $\m \phi$ and $\m \eta$ to obtain
\begin{align}
p(\boldsymbol{A},\boldsymbol{z}|\alpha,\gamma,\kappa,\lambda) 
& = \iint\! \! d\! \boldsymbol{\eta} d\! \boldsymbol{\phi}\  p(\boldsymbol{A}|\boldsymbol{\theta},\boldsymbol{\eta},\boldsymbol{z})
p(\boldsymbol{\eta}|\kappa,\lambda)
p(\boldsymbol{\phi}|\boldsymbol{z},\gamma)p(\boldsymbol{z}|\alpha) \nonumber \\
& = \frac{1}{\prod_{i\leq j} A_{ij}!\prod_i 2^{A_{ii}}}
 \prod_{\ell\leq m} \frac{G\left(N^+_{\ell m} + \kappa, N_{\ell m} + \lambda\right)}
{G(\kappa, \lambda)} \nonumber \\
& \times \left[\prod_\ell
  \frac{B\left(\gamma \boldsymbol{1}_{(n_\ell)} + (\hat{k}_i)_{i:z_i=\ell}\right)}{B(\gamma \boldsymbol{1}_{(n_\ell)})} n_\ell^{\hat{k}_\ell}
  \right] \left[\frac{\alpha^L\Gamma(\alpha)}{\Gamma(N+\alpha)}\prod_{\ell=1}^L\Gamma(n_\ell) \right].  \label{eq:15}
\end{align}
In the above derivation we exploit that $\sum_{z_i=\ell} \theta_i=n_\ell$ and thus the derivation requires access to the entire network. As a result, the inference of our generative model is reduced to determining the posterior distribution of the assignment of nodes to groups, $\m z$.

The assignment matrix $\m z$ is inferred using standard Gibbs sampling~\cite{kemp2006learning}, and using the Bayesian framework we can treat the hyperparameters $\gamma,\alpha,\lambda$ and $\kappa$ as random variables. In particular, we will invoke the non-informative prior $p(x) \propto x^{-1}$ for all four parameters and infer them using random-walk Metropolis updates of the form $x^* = \exp(\log x + z)$, $z \sim N(0,\sigma=0.1)$. For each Gibbs sweep over $\m z$, we performed 20 Metropolis-Hastings updates of the hyperparameters. While Metropolis-Hastings with random proposals is not very computational efficient, we noticed throughout the experiments this step had a small computational cost compared to sampling $\m z$.

\subsection{Imputation and link prediction}
Missing (unobserved) links commonly occur in network and predicting missing links is an important goal of network modelling. Comparing the prediction of a model on unobserved data to the actual value is furthermore a popular way to validate a model. In addition the self-links $A_{ii}$ are often unknown or, if the network cannot contain self-links such as the case of a friendship network, they should be treated as axillary variables that are integrated out.

For the IDCSBM the (marginalized) expression for $\m z$ in eq.~\eqref{eq:15} requires access to all entries in the adjacency matrix and so it is not possible to marginalize over missing data simply by ignoring the corresponding terms in the likelihood function. To overcome this difficulty we marginalize over missing entries by formulating a Markov chain Monte Carlo algorithm jointly over the parameters and the missing links. This is done by sampling $\m z$ and the hyperparameters using Gibbs sampling and random-walk Metropolis Hastings, and then conditionally on $\m A$ and $\m z$ drawing values of $\eta_{\ell m}$ and $(\phi_i)_{i}$ conditional on the full matrix $\m A$ and assignments $\m z$ and conditionally on these values draw the values of $\m A$ corresponding to the missing links from the Poisson distribution eq.~\eqref{eq:5}. This corresponds to imputing the missing values from their predictive distribution in each step of the MCMC algorithm and, assuming convergence of the Markov chain, is equivalent to marginalizing out the missing links. We use this framework both to handle self-links but also for link prediction in general. Another popular method to predict missing data is simply replacing missing entries of $\m A$ with $0$~\cite{karrer2011stochastic,Guimera14122009,clauset2008hierarchical}, however as the diagonal of $\m A$ is often fully missing, and the poisson rate for $A_{ii}$ is proportional to $\theta_i^2$, this approach would create an undesirable bias for $\theta_i$.

\subsection{Properties of the model} \label{sec:properties}
An important property of the model is that it can accurately learn the degree distribution of the data and the link-density between the groups. Suppose $\m A_0$ is an observed network and let $\m z$ be any fixed cluster. Conditional on $\m A_0$ and $\m z$ we may compute the posterior over $\m \eta$, $\m \theta$ and check if these distributions accurately reflect relevant properties of $\m A_0$. First notice from eq.~\eqref{eq:11} the posterior distributions of $\m \eta, \m \theta$ are
\begin{align}
p(\eta_{\ell m} | \m A_0, \m z) & = \Gam(\eta_{\ell m}\  |\  N^+_{\ell m} + \kappa, N_{\ell m} + \lambda) \\
p\left(\left(\frac{\theta_i}{n_\ell}\right)_{z_i = \ell}  | \m A_0, \m z\right) & = \Dir\left(\left(\frac{\theta_i}{n_\ell}\right)_{z_i = \ell} \ | \ \gamma \m 1_{n_\ell} + (\hat{k}_{i})_{z_i = \ell} \right)
\end{align}
Recall for two Poisson distributed random variables $X \sim \Poisson(a)$, $Y \sim \Poisson(b)$ their sum is Poisson with rate $a+b$: $X + Y \sim \Poisson(a+b)$. This, along with the derivation eq.~\eqref{eq:11}, allow us to compute the various properties of the model.

First consider the total interaction strength between two groups $\ell$ and $m$. The interaction $\sum_{i \leq j} \delta_{z_i = \ell}\delta_{z_j = m} A_{ij}$, considered as a random variable, is then distributed as $\Poisson(\eta_{\ell m}N_{\ell m})$. If $X \sim \Poisson(\lambda)$ then $\mathbb{E}[X] = \lambda$ and so the average between-group interaction is (the expectation is with respect to $p(\cdot | \m A_0,\m z)$)
\begin{align}
\mathbb{E}\left[  \sum_{i \leq j} \delta_{z_i = \ell}\delta_{z_j = m} A_{ij} \right] & = \mathbb{E}\left[ N_{\ell m} \eta_{\ell m} \right]  = \frac{N_{\ell m}}{N_{\ell m} + \lambda}( N^+_{\ell m}+ \kappa ).\label{eq:17}
\end{align}
For analytical simplicity, we will consider the degree plus the diagonal element. To this end define the degree of node $i$ as $d_i = \sum_{j} A_{ij}+A_{ii}$. Since each $A_{ij}$ is Poisson distributed the degree too is a Poisson random variable. If $z_i = \ell$ then $d_i$'s distribution is given by
\begin{align}
d_i & \sim \Poisson\left(\sum_{j \neq i} \theta_i \eta_{\ell z_j} \theta_j + 2\frac{1}{2}\theta_i^2 \eta_{\ell \ell} \right) = \Poisson\left( \theta_i \sum_{m} \eta_{\ell m}n_m  \right).
\end{align}
We may now compute the average, again with respect to $\m A_0$ and fixed $\m z$:
\begin{align}
\mathbb{E}[d_i] & = \mathbb{E}\left[ \theta_i \sum_{m} \eta_{\ell m}n_m   \right] \nonumber  \\
& =  n_\ell \frac{\hat{k}_i + \gamma}{\sum_{j : z_j = \ell } \hat{k}_j  + \gamma n_\ell }\sum_{m} \frac{N^+_{\ell m}+ \kappa}{N_{\ell m} + \lambda}n_m \nonumber \\
& = (\hat{k}_i + \gamma) \sum_m \frac{N_{\ell m} 2^{\delta_{\ell m}} }{N_{\ell m} + \lambda} \frac{N^+_{\ell m} + \kappa}{\sum_{h} N^+_{\ell h}2^{\delta_{\ell h}} + \gamma n_\ell}.\label{eq:22}
\end{align}
Assuming the groups are fairly large, and in the low limit of the prior $\gamma$, the sum will be $1$ to first order. The derivations eq.~\eqref{eq:17} and \eqref{eq:22} show in the limit of large systems the relative influence of the prior terms will vanish and the model will accurately capture the between-group link density as well as the node degree.

\section{Results and Discussions}
We analyze synthetic datasets generated from our model as well as seven real networks from the literature.

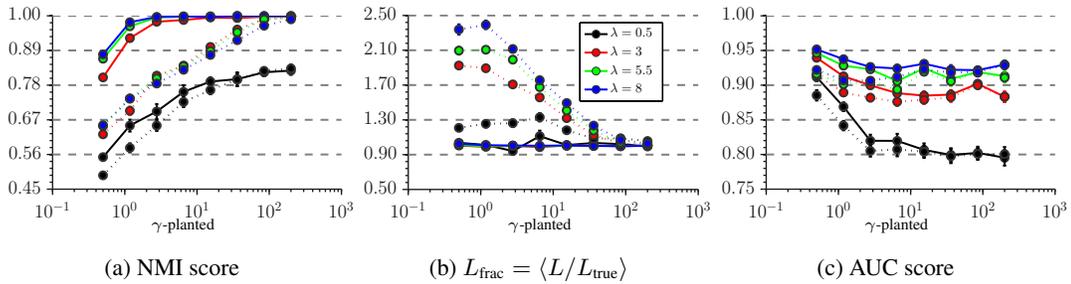
\begin{figure}
        \centering
        ~ 
        \begin{subfigure}[b]{0.33\textwidth}
        \begin{tikzpicture}
                    \node (n1) {\includegraphics[width=1.0\textwidth]{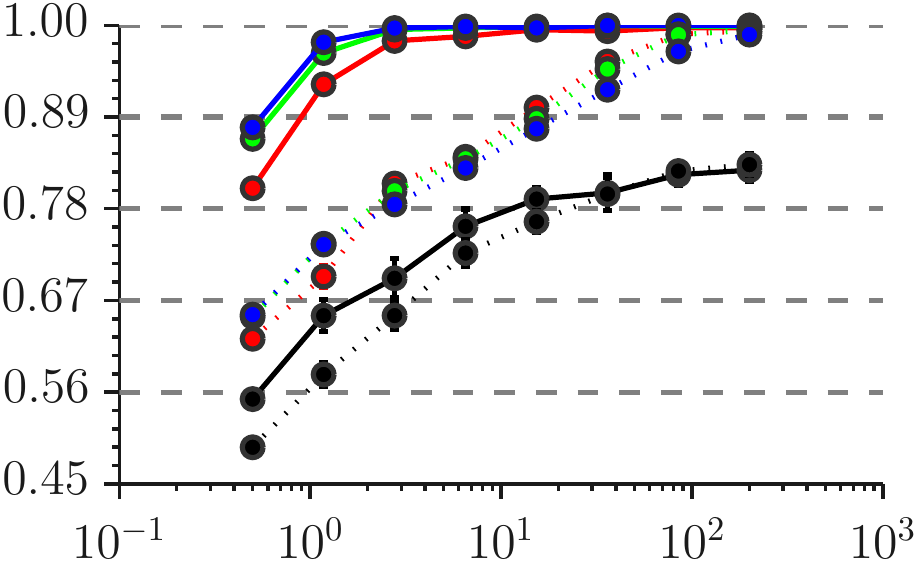}};
                    \node[below of=n1,yshift=-0.5cm] {\tiny{ $\gamma$-planted} };
                \end{tikzpicture}
                \caption{NMI score}
                \label{fig:1b}
        \end{subfigure}~
         \begin{subfigure}[b]{0.33\textwidth}
         \begin{tikzpicture}
                    \node (n1) {\includegraphics[width=1.0\textwidth]{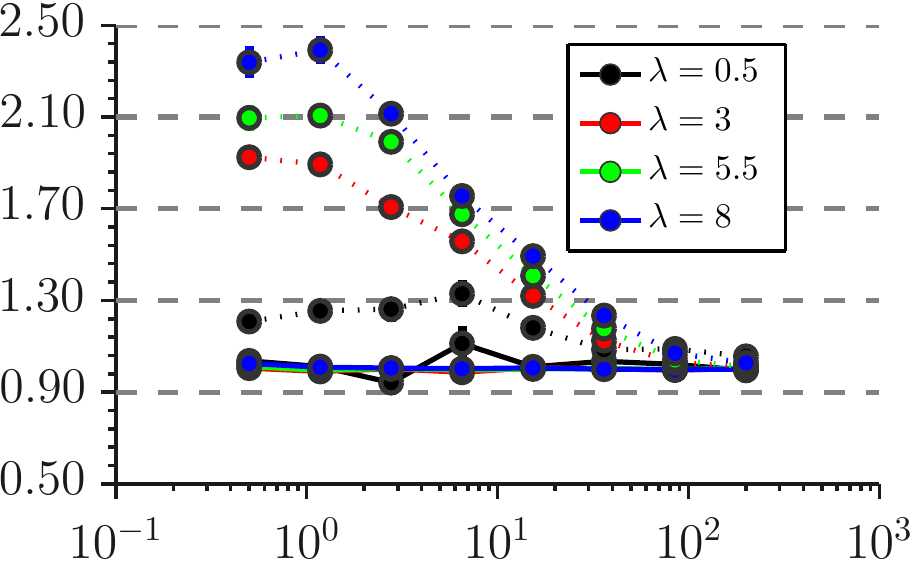}};
                    \node[below of=n1,yshift=-0.5cm] {\tiny{ $\gamma$-planted} };
                \end{tikzpicture}

                \caption{$\Lfrac = \left\langle L / L_\text{true} \right \rangle$}
                \label{fig:1c}
        \end{subfigure}~
        \begin{subfigure}[b]{0.33\textwidth}
                \begin{tikzpicture}
                    \node (n1) {\includegraphics[width=1.0\textwidth]{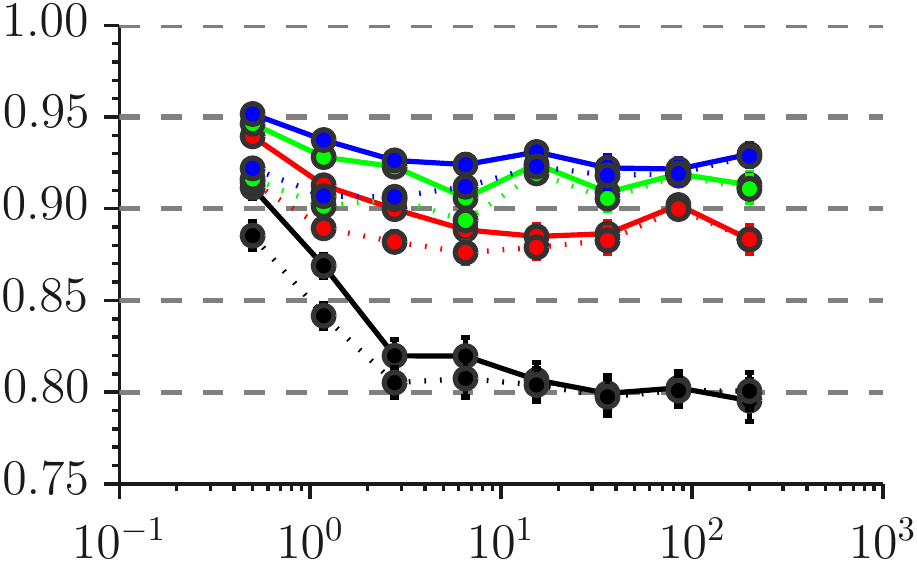}};
                    \node[below of=n1,yshift=-0.5cm] {\tiny{ $\gamma$-planted} };
                \end{tikzpicture}
                \caption{AUC score}
                \label{fig:1a}
        \end{subfigure}%
        \caption{IDCSBM and ISBM results on simulated network data. The plots show the normalized mutual information (NMI), the ratio of estimated to true number of components $\Lfrac$ as well as the area under curve (AUC) of the receiver operator characteristics as computed by running the proposed methods on networks produced from the generative model of the IDCSBM with different values of $\lambda$ and $\gamma$. The fully drawn lines indicate results for the IDCSBM, dotted lines indicate results for ISBM.}\label{fig:1}
\end{figure}

\subsection{Synthetic data}
In our synthetic simulation studies we generated networks of $N=80$ nodes from our generative model with the parameters $\kappa$ and $\alpha$ fixed at $\kappa=0.5$ and $\alpha=4$ and under different values of $\lambda$ and $\gamma$.

Each such network was analyzed using out Infinite Degree Corrected Stochastic Block Model (IDCSBM) as well as the corresponding infinite SBM (ISBM) without degree correction. In figure \ref{fig:1} the normalized mutual information (NMI), the ratio of true number of components to estimated number of components $\Lfrac=\langle \frac{ L }{L_{\text{true}}}\rangle$ and the area under curve (AUC) of the receiver operator characteristic are given (error bars indicate standard deviation of the mean where the deviation is computed over 10 restarts of the sampler). In the analysis we ran the samplers for $1000$ iterations and discarded the first half as burnin. The AUC scores were computed by treating 5\% of the links and a similar number of non-links as missing.

From the plot of the NMIs we see that the degree corrected model (IDCSBM) better recover the true generated group structure than the uncorrected model (ISBM) and as expected the performance of the two methods converge as $\gamma$ increases corresponding to networks which does not exhibit degree heterogeneity. Furthermore, the IDCSBM recover the correct number of groups whereas the ISBM generates more than the true number of groups in order to account for the effect of a skewed degree distribution. The predictive performance as quantified by the AUC scores are more or less similar with a tendency of slightly better predictions for the IDCSBM. As expected this is most notable for small values of $\gamma$. We further observe that structure is better recovered when the contrast in the interactions are high as influenced by the values of $\lambda$. This too can be expected since very sparse networks presumably has little recoverable structure.

\subsection{Real data}
We analyzed the following seven networks
\begin{figure}
        \centering

        \begin{subfigure}[b]{0.5\textwidth}
                \begin{tikzpicture}
                    \node (n1) {\includegraphics[width=0.95\textwidth]{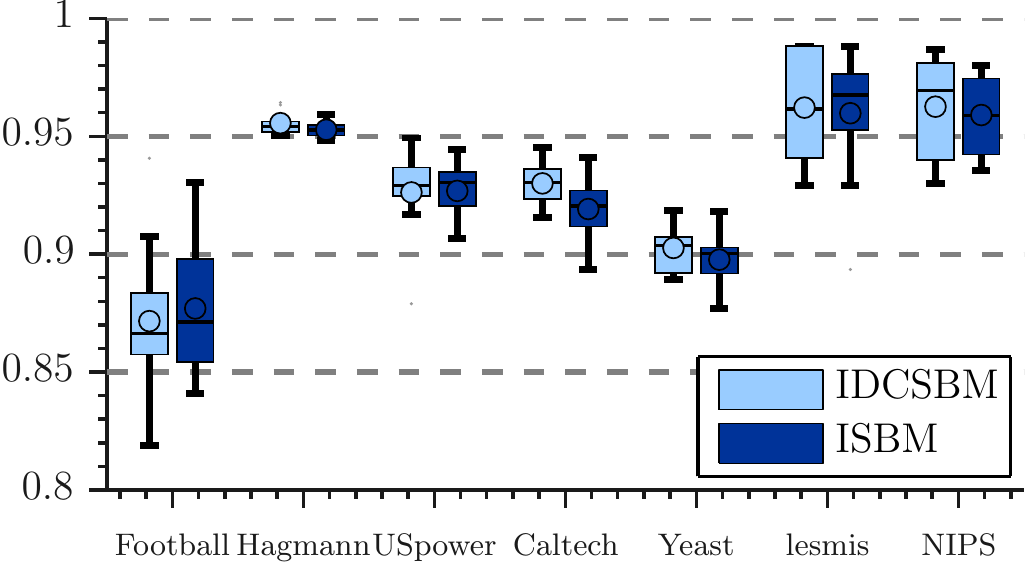}};
                    \node[left of=n1,rotate=90,yshift=2.5cm] {\tiny{ AUC score }};
                \end{tikzpicture}
                \caption{AUC score}
                \label{fig:2a}
        \end{subfigure}~
                \begin{subfigure}[b]{0.5\textwidth}
                \begin{tikzpicture}
                    \node (n1) {\includegraphics[width=0.95\textwidth]{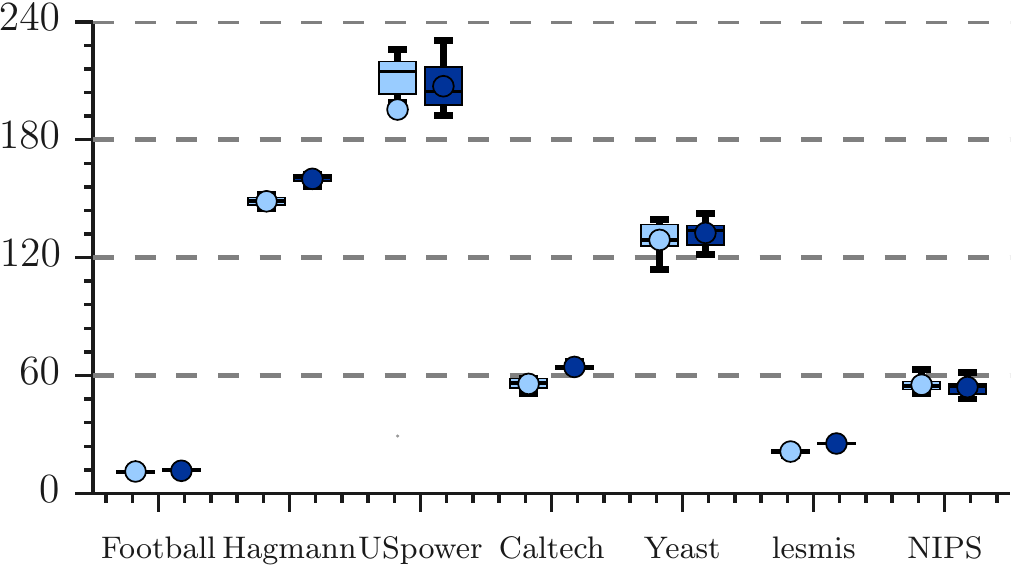}};
                    \node[left of=n1,rotate=90,yshift=2.5cm] {\tiny{ $\langle L \rangle$ }};
                \end{tikzpicture}
                \caption{Number of inferred Components}
                \label{fig:2K}
        \end{subfigure}%
        \caption{IDCSBM and ISBM results on the seven real network. To the left is given AUC scores and to the right the number of inferred groups $L$.}\label{fig:2}
\end{figure}
\begin{itemize}
\item \emph{Football:} Undirected unweighted network of American football games between 115 Division IA colleges in the Fall 2000 \cite{girvan2002community}.
\item \emph{Hagmann:} Undirected weighted network of the number of links between 998 brain regions as estimated by tractography from diffusion spectrum imaging across five subjects \cite{hagmann2008mapping}. I.e., the graph of each subject has been symmetrized, thresholded at zero and the five subject graphs added together.
\item \emph{USPower:} Undirected unweighted network of  4941 nodes representing the topology of the Western States Power Grid of the United States compiled by \cite{watts1998collective}
\item \emph{Caltech:} The Caltech39 social network from the Facebook100 dataset (available at \url{http://datahub.io/dataset/facebook100}).
\item \emph{Yeast:} The interaction network between 2361 proteins of yeast~\cite{bu2003topological}.
\item \emph{Lesmis:} Undirected and weighted graph of the co-appearences of 77 characters in Les Miserables by Victor Hugo \cite{knuth1993stanford}.
\item \emph{NIPS:} Undirected weighted network of the number of co-authorships between 234 authors of papers presented at the Neural Information Processing Systems 1-12 (available at \url{http://www.cs.nyu.edu/~roweis/data.html}).
\end{itemize}

\begin{figure}[h!]
        \centering
            \begin{tikzpicture}
                    \node (n1) {\includegraphics[width=0.5\textwidth]{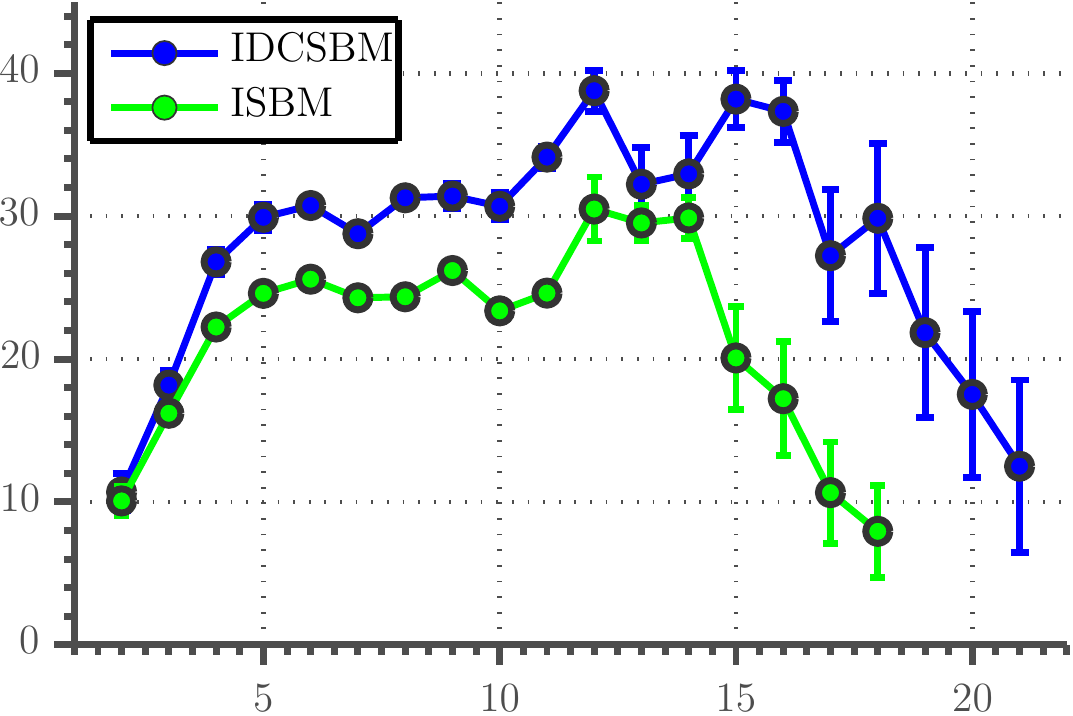}};
                    \node[left of=n1,rotate=90,yshift=2.8cm,xshift=0cm] {\small{  $\mathbb{E}[\textrm{std}[k_\ell]]$ }};
                    \node[below of=n1,yshift=-1.5cm] {\small{ Group size $n_\ell$ }};
                \end{tikzpicture}


        \caption{Variance of degree heterogeinity for the ISBM and IDCSBM for the Hagmann dataset. Each point $(k,y)$ is an estimate of the standard deviation of the degree distribution for nodes in a group $\ell$ of size $n_\ell = k$, see main text for details.}\label{fig:3}
\end{figure}

In figure \ref{fig:2} is shown the results for the IDCSBM and the ISBM on the seven networks in terms of AUC score treating 5\% of the links (and a similar number of non-links) as missing. Furthermore, the numbers of estimated components by the two models are given. The samplers were run for 1000 iterations (half discarded as burnin)
and the results are averaged over 10 restarts.

From figure~\ref{fig:2} it can be seen that in general the performance in predicting link as quantified by the AUC scores are on par for the IDCSBM and ISBM. However, as observed also in the synthetic study the IDCSBM model extracts less components than the ISBM for the Hagmann, Caltech, and Lesmis networks. Thus, the model allocates less groups when compared to the ISBM that allocates additional clusters in order to compensate for its lack of ability to explicitly account for degree.

Another way to examine this effect is to look at the degree distribution within each group. Since the groups have vastly different sizes it is hard to summarize this effect into a single number, however if we consider a fixed group structure $\m z$ and a single group $\ell$ of size $n_\ell$ we may compute the empirical mean $\mathbb{E}[k_\ell] = \frac{1}{n_\ell} \sum_{i : z_i = \ell} k_i$ and standard deviation $\mathrm{std}[k_\ell] = \sqrt{\frac{1}{n_\ell} \sum_{i : z_i = \ell} (k_i - \mathbb{E}[k_\ell])^2}$ of the degree within this group.

In figure~\ref{fig:3} we plotted the average of the empirical standard deviation of the degree distribution as a function of group size, that is, for each point $(k,y)$ in figure~\ref{fig:3}, $y$ is an estimate of $\mathbb{E}\left[ \mathrm{std}[k_\ell] \right]$
 where the expectation is conditional on $n_\ell = k$. This quantity is easily estimated based on the last $500$ states of a MCMC chain. The error bars are the standard deviation of the mean of each point based on 10 random restarts of the sampler.

 As can be seen, the IDCSBM discover larger groups of nodes confirming our previous findings in figure~\ref{fig:2} and, more importantly, the variance of the degree distribution within groups is larger than for the ISBM for all groups sizes. This show the compensation for degree heterogeneity not only affect a few large groups the IDCSBM lump together and the ISBM split apart, but groups of all sizes.


\begin{figure}
        \centering
    \begin{subfigure}[b]{0.5\textwidth}
    \begin{tikzpicture}
                    \node (n1) {\includegraphics[width=0.95\textwidth]{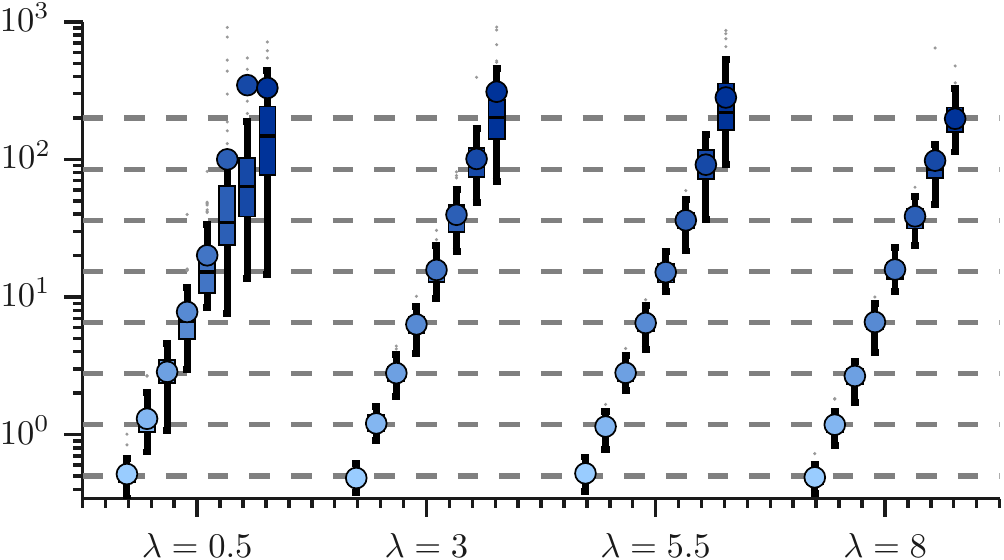}};
                    \node[left of=n1,rotate=90,yshift=2.5cm] {\tiny{ $\langle \gamma \rangle$ }};
                \end{tikzpicture}
                \caption{Recovery of $\gamma$ for artificial networks}
                \label{fig:gam1}
        \end{subfigure}~
        \begin{subfigure}[b]{0.5\textwidth}
         \begin{tikzpicture}
                    \node (n1) {\includegraphics[width=0.95\textwidth]{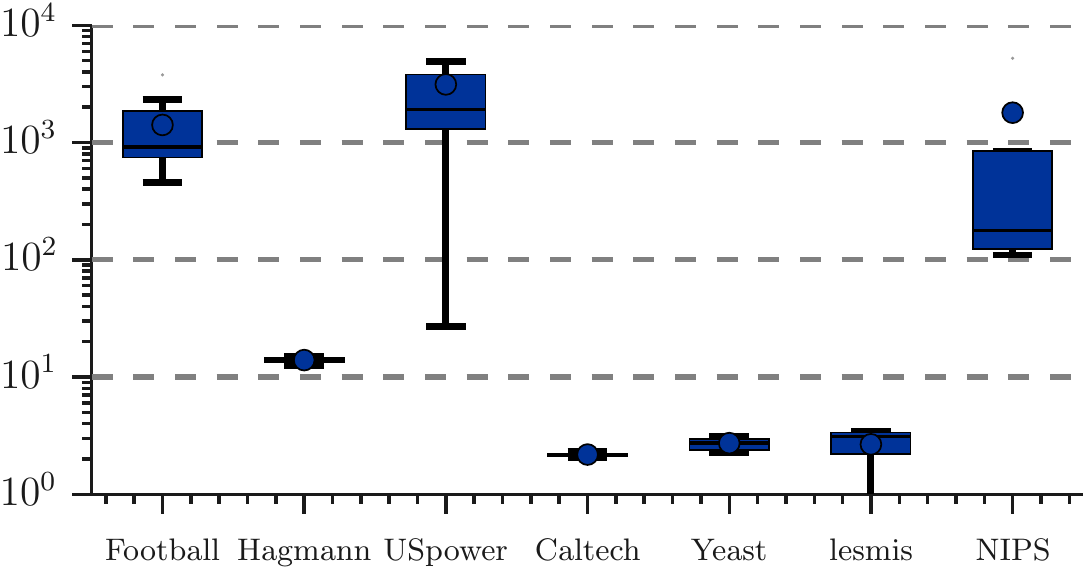}};
                    \node[left of=n1,rotate=90,yshift=2.5cm] {\tiny{ $\langle \gamma \rangle$ }};
                \end{tikzpicture}
        \caption{Inferred values of $\gamma$}\label{fig:gam2}
        \end{subfigure}
        \caption{Inferred values of $\langle \gamma \rangle $ for the real networks (right) and for the artificial networks (left). The box plots are created based on the mean of $\gamma$ for each of the 10 or 50 chains (real/artificial networks). For the artificial network, we group the networks according to the planted value of $\lambda$, and each box in a given group correspond to a particular planted value of $\gamma$. The horizontal lines indicate the planted values of $\gamma$. In the asymptotic limit of perfect sampling, the boxes should converge to points centered on the dotted lines.}\label{fig:gam}
\end{figure}

To better understand the role of $\gamma$, we examined the behaviour of the mean value of $\gamma$, $\langle \gamma \rangle$, across the random restarts of the chains both for the artificial and real datasets (see figure~\ref{fig:gam}). For the artificial datasets (figure~\ref{fig:gam1}) we grouped the networks according to the value of $\lambda$ and $\gamma$ used to generate the networks and plot the value of $\langle \gamma\rangle$ across the $50$ restarts. Consistent with the other findings, the model has more difficulties recovering the true value of $\gamma$ for very low link density ($\lambda = 0.5$) or when the planted value of $\gamma$ is very high, here 200 as the highest value. The later finding may be related to this value not being favoured by the prior. However the sampler generally recover the planted value of $\gamma$ well across chains.

For the real networks (figure \ref{fig:gam2}), the recovered values of $\langle \gamma \rangle$ across chains show quite high variability for some of the larger networks indicating they may exhibit mixing times significantly longer than the 1000 iterations used here. Notice that since high values of $\gamma$ is associated with a nearly vanishing effect of the degree, we see the model correctly identify the skewed degree distribution of the social network Caltech and Yeast, while indicating the effect of degree for the (very strongly) community-structured network Football and the spatially embedded USPower network is vanishing.

\section{Conclusion}

In this paper we extended the degree corrected stochastic block model (DCSBM)~\cite{karrer2011stochastic} to a non-parametric Bayesian generative model (the IDCSBM). The advantage of the proposed model being that the number of blocks, i.e. the distribution of the number of groups can be inferred, extending the model to an infinite representation similar to what has previously been done for the regular stochastic block model \cite{kemp2006learning,xu2006learning}. By exploiting the model is formulated generatively we have derived a Markov chain Monte Carlo algorithm which handle missing links explicitly by marginalize over missing entries. We have further shown we can learn the parameter $\gamma$ in the process and thereby determine the extent to which networks can use the degree correction parameter $\m \theta$ introduced in the degree corrected stochastic block model. We have shown analytically that under wide conditions the model will be able to accurately model between-group link density as well as node degree.

On synthetic and real networks we demonstrate that the IDCSBM can result in a more compact representation of network structure. The IDCSBM also tend to use fewer components than the ISBM while accounting equally well for the networks as quantified by the AUC link prediction scores. On synthetic data with degree-heterogeneity we have shown the proposed model, which correct for degree skewness, is able to infer the parameters controlling degree heterogeneity correct and obtain both a more compact and accurate representation. As expected, this also translate into improved link prediction. On real network data, we have shown a model which capture degree skewness does not dominate a model which does not in terms of link prediction, however the IDCSBM is able to consistently learn vastly different values of $\gamma$ and thereby the presence or absence of degree heterogeneity.

\bibliography{mm,ref2,references}
\bibliographystyle{plain}

\end{document}